\documentclass[11pt]{article}

\usepackage[margin=1in]{geometry}
\usepackage{amsmath}
\usepackage{amssymb}
\usepackage{booktabs}
\usepackage{graphicx}
\usepackage{hyperref}
\usepackage{microtype}
\usepackage[numbers,sort&compress]{natbib}

\title{Local Pheromone Network: Sparse Local Learning with Multi-Scale Synaptic Trails, Consolidation, and Replay}

\author{
Fu Xingcheng\\
\texttt{fxc@wlhex.com}
\and
Chen Xianjun\\
\texttt{xianjun.mx@gmail.com}
\and
Li Zhihao\\
\texttt{lizhihao@holdingfuture.com}
}

\date{June 2026}

\begin{document}
\maketitle

\begin{abstract}
Backpropagation-trained dense neural networks are powerful function approximators, but they couple learning across many parameters and can overwrite previous associations when tasks conflict. This paper describes Local Pheromone Network, a small research prototype for sparse, local, manually updated neural networks. In Local Pheromone Network, each output unit reads only a fixed local neighborhood of input units subject to geometric distance and molecular-tag compatibility. Each synapse stores a weight, a short-term pheromone trace, a long-term pheromone trace, and an optional consolidation state. Training does not call automatic differentiation. Instead, every layer performs a pheromone-weighted Hebbian-style update on a budgeted subset of local synapses selected from local error and co-activity. The update budget adapts online: it shrinks when loss improves and expands toward recently active neighborhoods when loss worsens. Optional mechanisms add structural plasticity, local replay, output masks for partitioned learning, and a target-free local contrastive step. We present the implementation, learning rule, and preliminary experiments on synthetic regression, partitioned memory, conflicting memory, consolidated conflict, structural plasticity, replay, and a synthetic long-context hybrid memory task. The prototype learns local linear rules, preserves partitioned memories through tags and masks, reduces forgetting under consolidation, and uses replay under conflict.
\end{abstract}


\section{Introduction}

Most contemporary neural networks use dense or structured differentiable modules trained by global backpropagation. This has produced remarkable results, but it also couples learning across wide parameter sets: a new task may adjust parameters needed by an older task, causing catastrophic forgetting \citep{mccloskey1989catastrophic,kirkpatrick2017overcoming}. Biological neural systems suggest a different collection of constraints: synapses are local, neuromodulatory signals alter plasticity, frequently used pathways become easier to reuse, and some memories become consolidated over time \citep{hebb1949organization,benna2016computational}.

Local Pheromone Network explores a simple computational analogue of those ideas. The model is not intended as a biologically faithful account of neural learning. Instead, it is an engineering prototype that asks: what happens if a neural layer stores only local sparse synapses, routes learning through pheromone-like traces, updates only a small budget of synapses, and protects mature high-pheromone connections from being overwritten?

The implementation studied here is contained in \texttt{local\_pheromone\_network.py}. It provides two main objects:
\begin{itemize}
    \item \texttt{LocalPheromoneLayer}, a sparse layer with tag-restricted local synapses and pheromone traces.
    \item \texttt{LocalPheromoneNetwork}, a stack of such layers trained by \texttt{local\_train\_step} rather than \texttt{loss.backward()}.
\end{itemize}

The central design choices are:
\begin{enumerate}
    \item each output unit connects to at most \texttt{max\_neighbors} compatible input units;
    \item compatibility is determined by spatial locality and integer tags;
    \item forward propagation uses a pheromone-gated effective weight;
    \item learning selects only a budgeted subset of synapses per output unit;
    \item short- and long-term pheromone traces evolve at different time scales;
    \item consolidated synapses become less plastic;
    \item replay can reapply old local examples when a new update appears to conflict with previous memory.
\end{enumerate}

The contributions of this manuscript are descriptive rather than benchmark-driven. We formalize the implemented model, document its mechanisms, and report preliminary behavior from the accompanying tests and evaluation scripts.

\section{Related Work}

The local update rule is inspired by Hebbian learning \citep{hebb1949organization}, but differs by using an explicit local error signal and a budgeted synapse-selection mechanism. Associative memory has a long history in neural computation, including Hopfield networks \citep{hopfield1982neural}. Local Pheromone Network models are not energy-based attractor networks, but they share the goal of storing persistent associations in weights.

The consolidation mechanism is related in spirit to complementary learning systems and elastic synaptic models of memory preservation \citep{kirkpatrick2017overcoming,benna2016computational}. However, Local Pheromone Network consolidation is per-local-synapse and driven by pheromone maturity, not Fisher information or gradient-derived parameter importance. Local replay follows the general idea that old examples can reduce forgetting, but uses the same manual local update rule rather than gradient replay.

The model also touches structural plasticity and sparse neural computation. Instead of training a dense matrix and later pruning, the layer is sparse by construction: each output unit stores only a local list of candidate synapses. Optional structural plasticity can prune low-pheromone low-consolidation synapses and sprout tag-compatible new synapses in active neighborhoods.

\section{Model}

\subsection{Units, Geometry, and Tags}

Let a layer receive an input vector $x \in \mathbb{R}^{d_\mathrm{in}}$ and produce an output $y \in \mathbb{R}^{d_\mathrm{out}}$. Each input and output unit has a normalized position in a one- or two-dimensional geometry. If a shape is provided, such as $(r,c)$, units are placed on a grid; otherwise they are placed on a one-dimensional line. Each unit also has an integer tag.

An input unit $i$ may be considered as a candidate for output unit $j$ only if
\begin{equation}
    |t^\mathrm{out}_j - t^\mathrm{in}_i| \leq \Delta_t,
\end{equation}
where $\Delta_t$ is \texttt{tag\_distance}. If a connection radius is provided, the candidate should also lie within that geometric radius. The implementation falls back to nearest tag-compatible inputs if no candidate exists inside the radius, but it does not fall back to tag-incompatible inputs.

For each output $j$, the layer selects at most $K=\texttt{max\_neighbors}$ nearest compatible inputs. These indices are stored in
\begin{equation}
    N_{j,k} \in \{1,\ldots,d_\mathrm{in}\},
\end{equation}
with a boolean mask $M_{j,k}$ indicating valid slots. The trainable state is therefore sparse:
\begin{equation}
    W \in \mathbb{R}^{d_\mathrm{out} \times K},
\end{equation}
rather than a dense $d_\mathrm{out}\times d_\mathrm{in}$ matrix.

\subsection{Pheromone-Gated Forward Pass}

Each synapse has a short-term pheromone $s_{j,k}$, a long-term pheromone $p_{j,k}$, and a consolidation value $c_{j,k}$. The effective pheromone is
\begin{equation}
    \phi_{j,k}
    =
    \frac{\alpha_s s_{j,k} + \alpha_p p_{j,k}}
         {\alpha_s+\alpha_p},
\end{equation}
when $\alpha_s+\alpha_p>0$, otherwise it defaults to the long-term pheromone. In the implementation these coefficients are \texttt{short\_pheromone\_weight} and \texttt{long\_pheromone\_weight}.

The forward pass first gathers local inputs
\begin{equation}
    \tilde{x}_{j,k} = x_{N_{j,k}}.
\end{equation}
The pheromone gate normalizes pheromone mass per output row:
\begin{equation}
    g_{j,k} = 0.5 + 0.5 \cdot
    \mathrm{clip}\left(
        \frac{\phi_{j,k}}{\sum_{k'} \phi_{j,k'} + \epsilon}
        \sum_{k'} M_{j,k'},
        0,2
    \right).
\end{equation}
The layer output is then
\begin{equation}
    y_j = b_j + \sum_{k=1}^{K} M_{j,k} \, W_{j,k} \, g_{j,k} \, \tilde{x}_{j,k}.
\end{equation}
Thus pheromone does not replace weights; it modulates which local pathways matter more during inference.

\subsection{Network Stack}

A \texttt{LocalPheromoneNetwork} stacks one or more layers:
\begin{equation}
    h^{(0)} = x,\quad
    z^{(\ell)} = L^{(\ell)}(h^{(\ell-1)}),\quad
    h^{(\ell)} = \sigma(z^{(\ell)})
\end{equation}
for hidden layers, with a linear final output. Supported activations are \texttt{tanh}, \texttt{relu}, \texttt{sigmoid}, \texttt{gelu}, and \texttt{identity}. All parameters have \texttt{requires\_grad=False}; learning is manual.

\section{Local Learning Rule}

\subsection{Loss and Output Mask}

The supervised local update uses mean squared error. An optional output mask $m$ restricts the active output region:
\begin{equation}
    \mathcal{L} =
    \frac{\sum_i m_i(\hat{y}_i-y_i)^2}{\sum_i m_i}.
\end{equation}
This is useful for partitioned memory: a task can update only its assigned output region instead of treating unrelated regions as zero targets.

\subsection{Budget Adaptation}

The network maintains a current update budget $B$, the maximum number of synapses per output unit to update. Let $\mathcal{L}_t$ be the current loss and $\mathcal{L}_{t-1}$ the previous loss. The mode is:
\begin{itemize}
    \item \textbf{warmup}: no previous loss exists;
    \item \textbf{exploit}: $\mathcal{L}_t < \mathcal{L}_{t-1}-\epsilon$; shrink $B$ by \texttt{shrink\_factor};
    \item \textbf{neighbor-follow}: $\mathcal{L}_t > \mathcal{L}_{t-1}+\epsilon$; grow $B$ by \texttt{grow\_factor};
    \item \textbf{steady}: otherwise.
\end{itemize}
The budget is clipped between configured minimum and maximum values.

\subsection{Layer Error and Synaptic Signal}

At the output, the local error is
\begin{equation}
    e = y-\hat{y}.
\end{equation}
For hidden layers the implementation propagates an approximate feedback signal through the effective local weights and activation derivatives. This is not automatic differentiation and not claimed to be a biologically exact signal; it is an engineering approximation that supplies each layer with a local error tensor.

For a layer, the synaptic signal is
\begin{equation}
    q_{j,k} = \mathrm{mean}_n \left(e_{n,j}\tilde{x}_{n,j,k}\right),
\end{equation}
clipped to \texttt{signal\_clip}. Co-activity is
\begin{equation}
    a_{j,k} =
    \mathrm{mean}_n
    \left(|\tilde{x}_{n,j,k}|\,|h_{n,j}|\right).
\end{equation}

\subsection{Synapse Selection}

Only a subset of synapses is updated. The base score is
\begin{equation}
    r_{j,k} = |q_{j,k}|(0.5+p_{j,k}).
\end{equation}
For each output unit, the top $B$ valid synapses are selected. In \textbf{neighbor-follow} mode, a bonus is added to synapses near recently updated synapses, encouraging local exploration around the last active frontier.

\subsection{Weight and Pheromone Update}

Selected weights are updated as
\begin{equation}
    W_{j,k} \leftarrow
    W_{j,k} + \eta \, q_{j,k} \, \mathbb{I}_{j,k} \, \rho_{j,k},
\end{equation}
where $\mathbb{I}_{j,k}$ is the selection mask and $\rho_{j,k}$ is a plasticity multiplier derived from consolidation. Unselected active connections may decay by \texttt{synapse\_decay}. Weights and bias are clipped.

Pheromone reinforcement is derived from $a_{j,k}+|q_{j,k}|$, with a mode-dependent multiplier. Short- and long-term pheromones evaporate at different rates:
\begin{align}
    s_{j,k} &\leftarrow (1-\lambda_s \rho_{j,k})s_{j,k}
        + \gamma_s R_{j,k}\mathbb{I}_{j,k},\\
    p_{j,k} &\leftarrow (1-\lambda_p \rho_{j,k})p_{j,k}
        + \gamma_p R_{j,k}\mathbb{I}_{j,k}.
\end{align}
Both traces are clipped to configured pheromone bounds.

\subsection{Consolidation}

Consolidation is optional. When enabled, a synapse consolidates when the current loss is below a gate, the mode is exploit, the synapse is selected, reinforcement is positive, and long-term pheromone exceeds a threshold. Its growth is proportional to mature pheromone:
\begin{equation}
    c_{j,k} \leftarrow
    \mathrm{clip}_{[0,1]}
    \left(
        c_{j,k}(1-\delta_c) +
        \gamma_c R_{j,k}\mathbb{I}_{j,k}
        \frac{\max(p_{j,k}-\theta_c,0)}{p_\mathrm{max}-\theta_c}
    \right).
\end{equation}
Plasticity is reduced by consolidation:
\begin{equation}
    \rho_{j,k}
    =
    \max(\rho_\mathrm{floor}, 1-\beta c_{j,k}).
\end{equation}
This protects old memory but slows adaptation on directly conflicting tasks.

\subsection{Structural Plasticity}

If enabled, structural plasticity prunes low-pheromone, low-consolidation, unselected synapses. Empty slots can sprout new synapses to tag-compatible candidates near active inputs. Sprouting is still bounded by \texttt{max\_neighbors}; the layer changes which local inputs it reads, not the total number of slots.

\subsection{Replay}

When \texttt{replay\_capacity} is positive, low-loss examples from exploit or steady mode can be stored. If a later loss exceeds the previous loss by \texttt{replay\_trigger\_margin}, stored examples are replayed by calling the same local update rule with \texttt{allow\_replay=False}. Replay is local and does not invoke backpropagation.

\paragraph{Training step sketch.}
\begin{enumerate}
    \item Compute prediction and layer caches.
    \item Compute MSE loss, optionally masked by output region.
    \item Adapt update budget and choose mode from loss trend.
    \item Compute output error and approximate per-layer local errors.
    \item For each layer, gather local pre-synaptic activities; compute local signal $q$ and co-activity $a$; select top-budget synapses using $|q|(0.5+p)$ and optional frontier bonus; update selected weights; evaporate and reinforce pheromones; and update consolidation plus optional structural plasticity.
    \item Store a replay candidate if loss and mode pass gates.
    \item If conflict is detected, replay stored examples.
    \item Return loss, mode, active synapses, budget, and replay count.
\end{enumerate}

\section{Experimental Setup}

Experiments are lightweight unit and prototype evaluations from the repository. They should be read as behavioral probes, not as a competitive benchmark. Unless otherwise noted, the model uses PyTorch tensors but does not call \texttt{loss.backward()} for Local Pheromone Network parameters.

\subsection{Synthetic Local Regression}

The base task maps 12 inputs to 3 outputs, each output depending on a different local group of 4 inputs. Tags constrain each output to its corresponding input group. After 80 local update steps, mean-squared error decreases from 1.171949 to 0.008426.

\subsection{Partitioned Memory and Conflict}

Two memory settings are tested. In the partitioned probe, task A uses output region 0 and input tag 0, while task B uses output region 1 and input tag 1. Output masks ensure only the active task region receives error. In the conflict probe, task B uses the same output region and local synapses as task A but with the opposite target.

\subsection{Consolidated Conflict}

The consolidated variant first trains task A until low loss, then trains the conflicting task B. Strong consolidation settings are used deliberately to test whether mature synapses resist overwriting.

\subsection{Extension Probes}

Additional tests check structural plasticity, multi-scale pheromone decay, replay under conflict, and local contrastive learning.

\subsection{Hybrid Sliding Model}

The hybrid model combines a conventional sliding convolutional predictor with a local pheromone window memory:
\begin{equation}
    \mathrm{logits}_{hybrid}
    =
    \mathrm{logits}_{conv}
    +
    \lambda \,\mathrm{logits}_{memory}.
\end{equation}
The convolutional branch is trained by gradient descent; the memory branch uses Local Pheromone Network updates. The reported probe is a synthetic long-context rule in which the correct output depends on information outside the short convolutional receptive field.

\section{Results}

\subsection{Synthetic Local Regression}

\begin{table}[h]
\centering
\caption{Synthetic local regression with tag-compatible local neighborhoods.}
\begin{tabular}{lrr}
\toprule
Metric & Before & After \\
\midrule
MSE & 1.171949 & 0.008426 \\
Update mode & -- & exploit \\
Budget per output & -- & 1 \\
Active synapses & -- & 3 \\
\bottomrule
\end{tabular}
\end{table}

The result confirms that the local update rule can fit a simple local linear mapping when the geometry and tags match the task structure.

\subsection{Long-Term Memory}

\begin{table}[h]
\centering
\caption{Memory probes. Lower loss is better.}
\begin{tabular}{lrrrr}
\toprule
Probe & A initial & A learned & A recall after B & Ratio \\
\midrule
Partitioned & 1.320694 & 0.017469 & 0.018697 & 1.07x \\
Conflicting & 1.988494 & 0.023737 & 2.198060 & 92.60x \\
\bottomrule
\end{tabular}
\end{table}

Partitioned memories are preserved because tasks occupy different tag/output regions. Directly conflicting memories overwrite each other, as expected.

\begin{table}[h]
\centering
\caption{Consolidated conflict comparison.}
\begin{tabular}{lrrrr}
\toprule
Model & A learned & A recall after B & B learned & Forgetting ratio \\
\midrule
Baseline & 0.001442 & 2.136281 & 0.020879 & 1481.87x \\
Consolidated & 0.004134 & 0.203638 & 2.127251 & 49.26x \\
\bottomrule
\end{tabular}
\end{table}

Consolidation substantially reduces forgetting, but at a high cost to learning the directly conflicting new task. This is the intended stability--plasticity trade-off.

\subsection{Extensions}

Structural plasticity changes the local neighborhood in a controlled test from connected inputs $[0,1,2]$ to $[0,1,5]$. Short-term pheromone shows rapid decay from 1.5470 to 0.0493, while long-term pheromone decays more slowly from 1.0069 to 0.8566. A replay conflict test enters \textbf{neighbor-follow} mode and triggers 2 replay steps. A local contrastive test increases the positive-negative margin from 0.000000 to 15.892842.

\subsection{Hybrid Memory Branch}

\begin{table}[h]
\centering
\caption{Hybrid sliding convolution plus local pheromone memory on a synthetic long-context rule.}
\begin{tabular}{lrr}
\toprule
Model & Accuracy & Mean gate \\
\midrule
Convolutional base & 0.08 & -- \\
Pheromone memory branch & 1.00 & -- \\
Additive hybrid & 1.00 & -- \\
Gated hybrid & 1.00 & 0.962 \\
\bottomrule
\end{tabular}
\end{table}

These results suggest Local Pheromone Network modules can be useful as local memory branches when a task benefits from partitioned long-context association. They do not establish broad sequence-modeling quality.

\section{Discussion}

\subsection{What the Prototype Demonstrates}

The experiments support five modest claims. First, a sparse tag-restricted local layer can fit simple local functions without autograd. Second, output masks and tags can separate memories across regions. Third, direct conflict on the same local synapses still causes forgetting. Fourth, consolidation reduces that forgetting by reducing plasticity of mature synapses. Fifth, local replay and structural plasticity can be integrated into the same local update loop.

\subsection{What It Does Not Demonstrate}

The model is not a competitive replacement for Transformer, recurrent, or convolutional networks. It does not solve large-scale sequence modeling. The current multi-layer feedback signal is an engineering approximation and should not be interpreted as a biologically exact local learning rule.

\subsection{Stability--Plasticity Trade-Off}

The consolidated conflict experiment makes the central trade-off visible. Strong consolidation protects previous memory by lowering plasticity, but it can block learning a directly contradictory new task. This is not a bug; it is the expected consequence of treating high-pheromone mature synapses as long-term memory. Practical systems need routing, partitioning, or controlled deconsolidation to handle true task conflicts.

\subsection{Dynamic Brain-Region Interpretation}

Tags, masks, and multiple Local Pheromone Network modules can be interpreted as crude brain-region abstractions. Strict tag distance isolates regions; larger tag distance allows cross-region communication. Output masks train only selected functional regions. Auto-partitioned wrappers can create separate Local Pheromone Network partitions when losses or conflicts suggest that a new context should not overwrite an old one.

\section{Limitations}

The implementation is a research prototype with several limitations:
\begin{itemize}
    \item input and output dimensions are fixed after construction;
    \item structural plasticity changes connections within fixed slots rather than growing unbounded capacity;
    \item hyperparameters strongly affect behavior;
    \item hidden-layer credit assignment remains approximate;
    \item no optimized sparse CUDA kernel is used;
    \item the replay buffer uses simple insertion order rather than learned retrieval;
    \item experiments are small and should be expanded before making strong empirical claims.
\end{itemize}

\section{Future Work}

Future work should add similarity-based replay retrieval, automatic input/output slot expansion, local uncertainty-aware routing, better hidden-layer credit assignment, visualization of pheromone and consolidation dynamics, and larger benchmarks for continual learning. Hybrid models are especially promising: a conventional fast learner can provide dense features, while Local Pheromone Network branches store local long-term associations and partitioned memories.

\section{Conclusion}

Local Pheromone Network is a compact prototype for sparse local learning with multi-scale synaptic traces. The model stores local weights, short- and long-term pheromones, and consolidation values; it updates only a budgeted subset of synapses without autograd. Preliminary experiments show that Local Pheromone Network models can learn simple local functions, preserve partitioned memories, reduce overwriting through consolidation, trigger replay under conflict, and improve a synthetic long-context predictor as a memory branch. The architecture provides a concrete testbed for local memory, brain-region partitioning, and stability--plasticity experiments.

\section*{Code Availability}

The reference implementation is the repository file \texttt{local\_pheromone\_network.py}. Related partitioning and hybrid adapters are:
\begin{itemize}
    \item \texttt{auto\_partitioned\_pheromone.py}
    \item \texttt{hybrid\_pheromone\_sliding\_model.py}
\end{itemize}
The main behavioral probes are the local regression, long-memory, brain-region, multitask, extension, and hybrid sliding tests in the repository.

\bibliographystyle{plainnat}
\bibliography{references}

\end{document}